\documentclass[10pt,twocolumn,letterpaper]{article}

\usepackage{cvpr}              

\usepackage{graphicx}
\usepackage{amsmath}
\usepackage{amssymb}
\usepackage{booktabs}

\usepackage[utf8]{inputenc} 
\usepackage{url}            
\usepackage{booktabs}       
\usepackage{amsfonts}       
\usepackage{nicefrac}       
\usepackage{microtype}      
\usepackage{lipsum}
\usepackage{xcolor}
\usepackage{multirow}
\usepackage{array}
\usepackage{pifont}
\usepackage{rotating} 
\usepackage{amssymb}
\usepackage{amsmath}
\usepackage{tabularx}
\usepackage{algorithm}
\usepackage{algorithmic}
\usepackage[switch]{lineno}
\graphicspath{ {./images/} }
\usepackage{listings}
\usepackage[accsupp]{axessibility}

\usepackage{xcolor}
\usepackage{colortbl}
\definecolor{Gray}{gray}{0.90}
\newcolumntype{g}{>{\columncolor{Gray}}c}
\definecolor{ffe1da}{RGB}{255,225,218}
\definecolor{F7E0D5}{RGB}{247,224,213}
\definecolor{darkF7E0D5}{RGB}{209,154,128}
\colorlet{Light}{F7E0D5}

\usepackage[pagebackref,breaklinks,colorlinks]{hyperref}

\usepackage{amsmath,amsfonts,bm}

\def\eqref#1{equation~\ref{#1}}

\def\1{\bm{1}}

\newcommand{\test}{\mathcal{D_{\mathrm{test}}}}

\def\rvu{{\mathbf{i}}}

\def\rvu{{\mathbf{u}}}
\def\rvv{{\mathbf{v}}}

\def\vc{{\bm{c}}}

\DeclareMathAlphabet{\mathsfit}{\encodingdefault}{\sfdefault}{m}{sl}
\SetMathAlphabet{\mathsfit}{bold}{\encodingdefault}{\sfdefault}{bx}{n}

\def\sR{{\mathbb{R}}}

\newcommand{\Amat}{{\bf A}}

\newcommand{\Cmat}{{\bf C}}
\newcommand{\Dmat}{{\bf D}}

\newcommand{\Pmat}{{\bf P}}
\newcommand{\Qmat}{{\bf Q}}

\newcommand{\Tmat}{{\bf T}}

\newcommand{\Zmat}{{\bf Z}}

\newcommand{\cv}{{\boldsymbol c}}

\newcommand{\pv}{{\boldsymbol p}}

\newcommand{\yv}{{\boldsymbol y}}

\newcommand{\zv}{{\boldsymbol z}}

\newcommand{\muv}{{\boldsymbol \mu}}
\newcommand{\nuv}{{\boldsymbol \nu}}

\newcommand{\Lcal}{\mathcal{L}}

\newcommand{\Rcal}{\mathcal{R}}

\newcommand{\Ep}{{\mathbb E}}

\usepackage[capitalize]{cleveref}
\crefname{section}{Sec.}{Secs.}
\Crefname{section}{Section}{Sections}
\Crefname{table}{Table}{Tables}
\crefname{table}{Tab.}{Tabs.}

\def\cvprPaperID{11488} 
\def\confName{CVPR}
\def\confYear{2022}

\begin{document}

\title{
Multi-modal Alignment using Representation Codebook 
}
\author{Jiali~Duan\textsuperscript{\rm 1}$^*$\quad~Liqun Chen\textsuperscript{\rm 1}\thanks{The first two authors contributed equally.}\quad~Son~Tran\textsuperscript{\rm 1}\quad~Jinyu Yang\textsuperscript{\rm 2} \\ 
Yi Xu\textsuperscript{\rm 1}\quad~Belinda Zeng\textsuperscript{\rm 1}\quad~Trishul Chilimbi\textsuperscript{\rm 1}\\
		{\textsuperscript{\rm 1} Amazon}
		~~
		{\textsuperscript{\rm 2} University of Texas at Arlington }
		 \\
		\small{\texttt{\{duajiali,liquchen,sontran,yxaamzn,zengb,trishulc\}@amazon.com}} \qquad
		\small{\texttt{\{jinyu.yang\}@mavs.uta.edu }} 
	}

\maketitle

\begin{abstract}

   Aligning signals from different modalities is an important step in vision-language representation learning as it affects the performance of later stages such as cross-modality fusion. Since image and text typically reside in different regions of the feature space, directly aligning them at instance level is challenging especially when features are still evolving during training.
   In this paper, we propose to align at a higher and more stable level using cluster representation. Specifically, we treat image and text as two ``views'' of the same entity, and encode them into a joint vision-language coding space spanned by a dictionary of cluster centers (codebook). We contrast positive and negative samples via their cluster assignments while simultaneously optimizing the cluster centers. To further smooth out the learning process, we adopt a teacher-student distillation paradigm, where the momentum teacher of one view guides the student learning of the other. We evaluated our approach on common vision language benchmarks and obtain new SoTA on zero-shot cross modality retrieval while being competitive on various other transfer tasks.
\end{abstract}


\vspace{-5mm}
\section{Introduction}

Vision language (V\&L) representation learning is the problem of learning a unified feature embedding using both image and text signals. Pretrained V\&L models have a great diversity of applications in various downstream tasks across different settings, e.g. via transfer learning~\cite{chen2020uniter,li2020oscar,zhang2021vinvl}. The main tasks in V\&L pretraining include aligning the feature spaces of different modalities (multi-modal alignment ~\cite{lu2019vilbert,chen2020uniter,li2020oscar,li2021align}) and capturing the interaction across modalities (cross-modal fusion, ~\cite{vaswani2017attention,dosovitskiy2020image}). Late fusion approaches such as CLIP\cite{radford2021learning} and ALIGN \cite{jia2021scaling} focused on the first task, while early fusion approaches such as OSCAR \cite{li2020oscar}, VinVL~\cite{zhang2021vinvl} and VilLT\cite{kim2021vilt} focused on the second one. In this work, we adopt a hybrid approach similar to ALBEF \cite{li2021align}, where features from image and text modalities were first aligned and then fused using a transformer encoder. The main focus of our work is on the feature alignment stage, which is challenging due to the fact that image and text inputs have very different characteristics. Existing approaches such as CLIP \cite{radford2021learning} and ALIGN \cite{jia2021scaling} have to rely on large training resources and on massive amount of data to obtain good alignments (400M and 1.8B image-text pairs respectively).

\begin{figure}[t]
  \centering
   \includegraphics[width=0.5\textwidth]{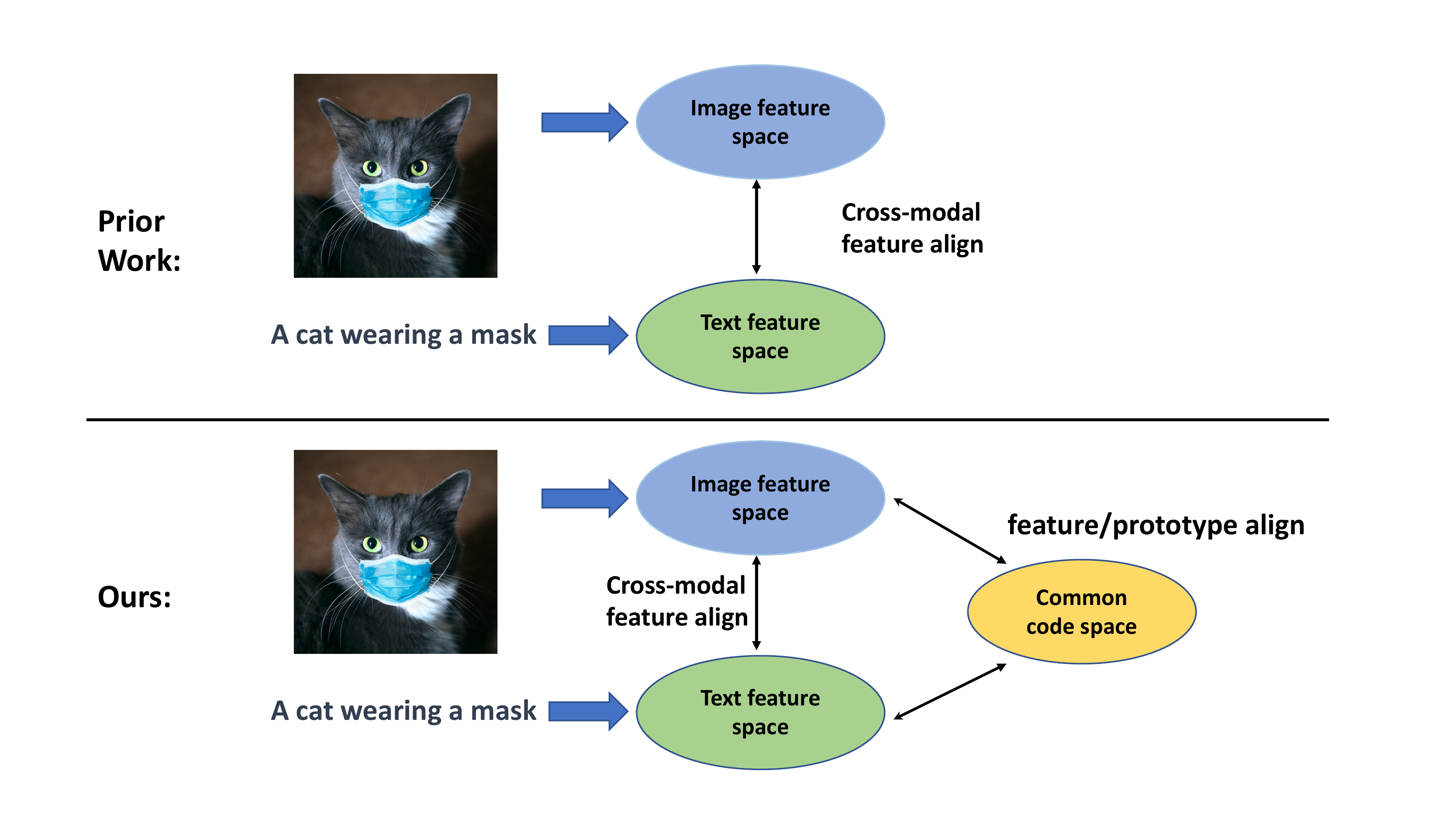}
   \caption{We propose to use a learnable codebook to better align the image and text modalities. The codebook serves as a ``bridge'' between the image and text features. Each codeword can be interpreted as a prototype, which enables contrasting image and text at the cluster level. We then solve an optimal transport~\cite{ambrosio2008gradient} problem to optimize the distance between each modality to the prototypes, which in turn optimizes the alignment between the two modalities. Prototype vectors are learned along with the feature encoders in our V\&L framework. }
   \label{fig:motivation}
\end{figure}

In this work, we propose a more efficient alignment strategy by using a codebook that quantizes the common text-image feature space into codewords. These codewords or cluster centers provide a more stable means for contrastive reasoning compared to individual text or visual features. We took the inspiration from SwAV\cite{caron2020unsupervised}, which was developed for self-supervised visual representation learning. In \cite{caron2020unsupervised}, two augmented versions (views) of the same input image were passed through a deep network for feature extraction. Visual embedding was learned by optimizing an objective function that enforces the consistency between the feature from one view and the assigned cluster from the other view. SwAV achieved impressive performance in various transfer tasks (see \cite{caron2020unsupervised}). Here, we carried out contrastive reasoning across modalities (image-text) instead of cross image views. Details are in Section \ref{sec: code}, but in a nutshell, we use a learnable codebook for both image and text modalities and train our model to predict the codeword assignment using either text or visual information. Effectively, visual and text features are lined up via aligning with the common codewords during training. See Figure \ref{fig:motivation} for an illustration.

The codebook can be considered as a quantized sample of the underlying output feature distribution. It is end-to-end learnable together with the model parameters. To avoid abrupt changes during training, we further employ momentum distillation, which has been widely used in previous self-supervised learning works such as BYOL~\cite{grill2020bootstrap}, DINO~\cite{caron2021emerging}, MoCo\cite{he2020momentum}. In brief, similar to ALBEF~\cite{li2021align}, for each of the image, text and fusion encoders, there is a corresponding encoder that is updated through moving average without gradient back propagation. These momentum encoders serve as teachers to guide the self-supervised learning process. Different from ALBEF~\cite{li2021align}, we use the teachers to guide codebook learning as well as for the cross-modal and intra-modal alignment.


The above two components are wired up to support the stable update of the codebook which, in turn, provides an efficient regularization mean for cross modality alignment. Experiment results (Section \ref{experiments}) show that our approach is competitive with state of the art across various benchmarks even when comparing with approach that use massive amount of data such as  CLIP \cite{radford2021learning} and ALIGN \cite{jia2021scaling}. 
In summary, our main contributions are as follows,
\begin{itemize}
    \item We propose a codebook-based approach for efficient vision-language alignment learning. It is an extension from self-supervised vision representation learning (SSL) to the multimodal setting.
    \item We introduce a new distillation algorithm that helps unimodal and crossmodal contrastive optimization as well as helps stablize codebook learning.
\end{itemize}

The rest of the paper is organized as follows. We introduce related work to ours in Section~\ref{sec:related}. In Section~\ref{sec:method}, we describe our framework, called \textbf{Co}debook Learning with \textbf{Dis}tillation (CODIS), and its two components, multimodal codebook learning and teacher-student distillation. Experimental results are presented in Section~\ref{experiments}. Section~\ref{sec:conclusion} concludes the paper. 
\section{Related Work}\label{sec:related}
\noindent\textbf{Vision-Language Pre-training (V\&L)}
V\&L pretraining is an active research area with many recent works. We review here the works that are most relevant to ours. 
Architecture wise, previous approaches can be broadly classified into two categories early fusion and late fusion. In early-fusion approaches ~\cite{su2019vl, kim2021vilt, chen2020uniter, li2020oscar}, image and text are transformed into sequences (tokenization) and passed to a single encoder (typically Transformer-based) for embedding generation. Thus multimodal signals are fused in the early stage. Whereas in late-fusion works ~\cite{radford2021learning,jia2021scaling}, separate encoders are used for image and text. Extracted features are typically fused during the later fine tuning stage. Our work is a hybrid between these two approaches, similar to~\cite{li2021align,yang2022vision}. The main difference is the codebook and various related contrastive losses.

In vision language learning, codebook has been used in a number of recent works, mostly for image tokenization. BEiT~\cite{bao2021beit} constructed a dictionary of visual words, then used it to form mask image modeling task in the same fashion as mask language modeling. SOHO~\cite{huang2021seeing} integrated visual dictionary to the main model and jointly trained both of them. Both works quantized the visual input space. In contrast, our codebook is used to quantize the joint output space
, where multimodal views are aligned via optimal transport~\cite{ambrosio2008gradient}.  
Other concurrent works to ours include~\cite{li2020unimo,li2021align}. They both align cross-modal instances using InfoNCE~\cite{oord2018representation}. In contrast, we enforce both unimodal and cross-modal alignment, both at the instance level and at the cluster level.

\noindent\textbf{Self-supervised Contrastive Learning}
The goal of contrastive learning~\cite{hadsell2006dimensionality} is to attract positive sample pairs and repulse the negative sample pairs. Recently, it has been widely used in computer vision for unsupervised, semi-supervised~\cite{duan2021slade} and self-supervised representation learning~\cite{he2020momentum,chen2020simple,caron2021emerging}. Contrastive reasoning is typically formed based on two augmented views of the same input image. One of the main challenge is feature collapsing, and in practice, a large number of negative samples are required, through either large batch size~\cite{chen2020simple} or memory banks~\cite{he2020momentum,wu2018unsupervised}, to alleviate this problem. Several recent works have shown that one can learn unsupervised features without discriminating instances. Deep clustering~\cite{caron2018deep} and SwAV~\cite{caron2020unsupervised} incorporate online clustering into Siamese networks. In BYOL~\cite{grill2020bootstrap}, features are trained by matching them to representations obtained by a momentum encoder. 
DINO~\cite{caron2021emerging} instantiates the momentum encoder with a vision-transformer and adopts a teacher-student distillation paradigm~\cite{hinton2015distilling,xie2020self,duan2021slade}. Our alignment techniques and momentum update were inspired by these works and can be considered as extensions to the multimodal setting.

\begin{figure*}[t]
  \centering
   \includegraphics[width=0.8\linewidth]{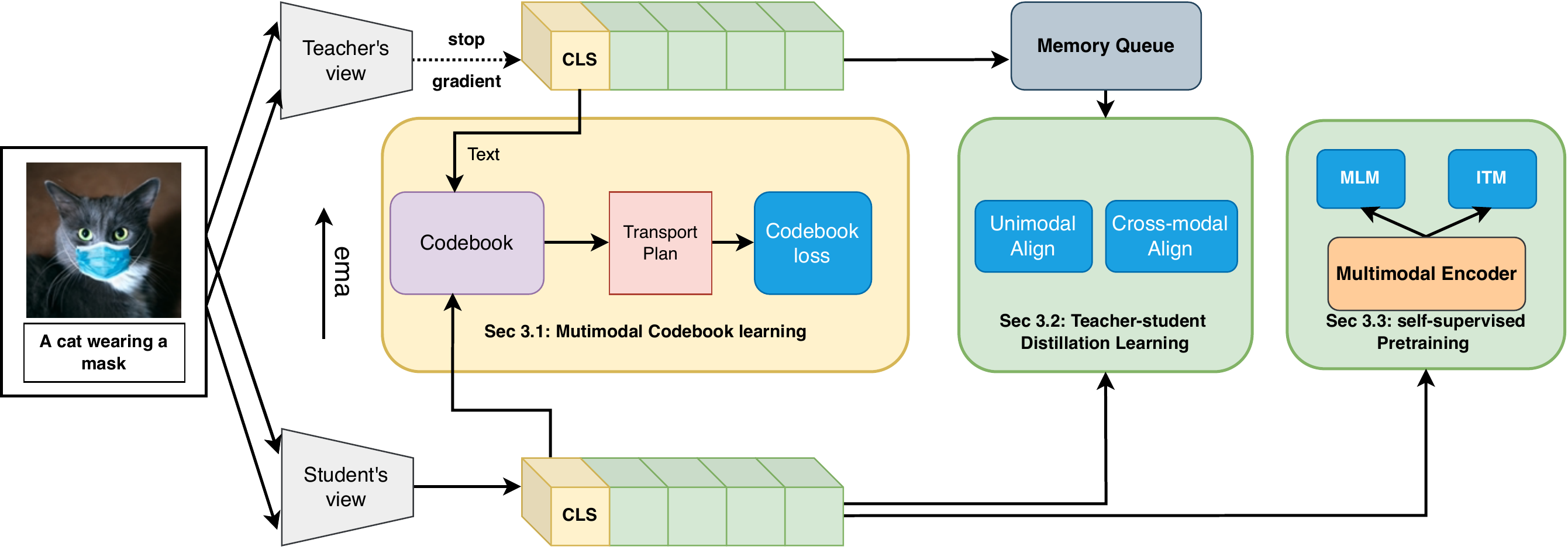}
   \caption{Overview of our framework. For simplicity, we only display a pair of teacher-student encoders (e.g., teacher for the image and student for the text) and similarly for the memory queue. The teacher is updated with an exponential moving average of the student (from the same modality). The codebook helps bridge the gap between the different modalities. The entire framework is end-to-end optimized.}. 
   \vspace{-3mm}
   \label{fig:framework}
\end{figure*}

\section{Method}\label{sec:method}
Our goal is to learn explicit alignment between image and text features to facilitate multimodal interactions. We illustrate CODIS in Figure~\ref{fig:framework} and propose a pseudo-code implementation in Algorithm~\ref{algo:frw-algo}. It shares some similarities with self-supervised contrastive learning~\cite{he2020momentum,caron2020unsupervised}. We treat image and text modalities as two views and adopt a teacher-student distillation paradigm~\cite{grill2020bootstrap, caron2021emerging} to enforce unimodal and cross-modal alignment. To overcome the gap between multimodal distributions, we also learn a codebook, which serves as a bridge to help align features between different modalities. We organize the content of this section as follows. 

In Section~\ref{sec: code}, we present multimodal codebook learning, how it's optimized and how to leverage it to resolve distribution mismatch between multimodal inputs. In Section~\ref{sec: itc}, we introduce how to achieve unimodal and cross-modal alignment under the teacher-student distillation learning formulation. Finally, we explain how our proposed two components integrate into the V\&L framework in Section~\ref{sec:pretrain}.

\begin{algorithm}[tb]
   \caption{CODIS pseudocode}
   \label{algo:DINO}
    \definecolor{codeblue}{rgb}{0.25,0.5,0.5}
    \lstset{
      basicstyle=\fontsize{7.2pt}{7.2pt}\ttfamily\bfseries,
      commentstyle=\fontsize{7.2pt}{7.2pt}\color{codeblue},
      keywordstyle=\fontsize{7.2pt}{7.2pt},
    }
\begin{lstlisting}[language=python]
# gs, gt: student/teacher networks for image 
# fs, ft: student/teacher networks for text 
# C: codebook d-by-K
# Qv, Qt: image/text queue, d-by-M
# tmp, learnable temperature
for (img, txt) in loader: # a minibatch with N samples
    # teacher/student's image view
    img_t, img_s = gt(img), gs(img) # N-by-d

    # teacher/student's text view
    txt_t, txt_s = ft(txt), fs(txt) # N-by-d
    
    # calculate codebook loss
    I2P, T2P = img_t@C, txt_t@C, # N-by-K
    Tg, Tf = IPOT(1-I2P), IPOT(1-T2P) # refer to Algo 2
    L_ot = Trace(I2P.t()@Tg).sum() + Trace(T2P.t()@Tf).sum()
    L_code = H(img_s@C, Tg) + H(txt_s@C, Tf) + L_ot
    
    # calculate alignment loss
    L_cross = H(img_s@Qt, img_t@Qt) + H(txt_s@Qv, txt_t@Qv)
    L_unimo = H(img_s@Qv, img_t@Qv) + H(txt_s@Qt, txt_t@Qt)
    L_align = L_cross + L_unimo
    
    # enqueue/dequeue
    update_queue(Qv, img_t, Qt, txt_t)
    
    # pretraining loss
    L_pretrain = L_itm + L_mlm
    
    loss = L_code + L_align + L_pretrain
    loss.backward() # back-propagate

    # student, teacher updates
    update(gs, fs) # SGD
    ema(gs, gt, fs, ft) # momemtum update
    

def H(s, t):
    t = t.detach() # stop gradient
    s = softmax(s / tmp, dim=1)
    return - (t * log(s)).sum(dim=1).mean()


\end{lstlisting}
\label{algo:frw-algo}
\end{algorithm}


\begin{algorithm}[!t]
\caption{IPOT Algorithm.}
\label{alg:ipot}
\begin{algorithmic}[1]
\STATE {\bfseries Input:} \footnotesize{
distance/similarity matrix $\Zmat$, $\Cmat$, $\epsilon$, probability vectors $\muv$, $\nuv$}
\STATE $\boldsymbol{\sigma}=\frac{1}{n}\mathbf{1_n}$, $\boldsymbol{\Tmat}^{(1)} = \mathbf{1} \mathbf{1}^\top$
\STATE $D_{ij} = d(\zv_i, \cv_j)$,$\Amat_{ij} = {\rm e}^{-\frac{\Dmat_{ij}}{\epsilon}}$
\FOR{$t=1,2,3\ldots$}
    \STATE $\Qmat = \Amat \odot \boldsymbol{\Tmat}^{(t)}$ \footnotesize{// $\odot$ is Hadamard product}
    \FOR{$k=1,2,3,\ldots K$}
        \STATE $\boldsymbol{\delta} = \frac{\muv}{n\Qmat{\boldsymbol{\sigma}}}$, $\boldsymbol{\sigma} = \frac{\nuv}{n\Qmat^\top\boldsymbol{\delta}}$
    \ENDFOR
    \STATE $\boldsymbol{\Tmat}^{(t+1)} = \text{diag}(\boldsymbol{\delta})\Qmat\text{diag}(\boldsymbol{\sigma})$
\ENDFOR
\STATE Return $\Tmat$

\end{algorithmic}
\end{algorithm}
\subsection{Multimodal Codebook Learning} \label{sec: code}

We propose to learn a codebook to facilitate aligning multimodal semantics. It's a collection of learnable prototypes or codewords. We use them interchangeably in this paper. With codebook, we encode image and text into a joint vision-language embedding space and learn the alignment by contrasting their prototype assignments. The codebook can also be interpreted as underlying feature distribution for the paired data~\cite{chen2020graph}. In this way, by aligning features from each modality with the codebook, we implicitly align multimodal features indirectly. In other words, the codebook serves as a ``bridge'' between the modalities (See Figure~\ref{fig:motivation}). 


We denote the learnable codebook as $\Cmat =\{\vc_1, \vc_2,\ldots, \vc_K\}\in \Rcal^{d_c \times K}$, where $d_c$ is the dimension for each code and $K$ equals to the number of codewords (i.e., $4$K). We set $d_c=256$, same as the dimension of projected image/text features. Each $\vc \in \Cmat$ is a prototype.

Given $N$ image or text feature vectors $\Zmat^m = [\zv^m_1, \ldots, \zv^m_N]$ (superscript $m$ denotes features extracted from the momentum teacher encoder), we compute an optimal cost mapping from the feature vectors to the prototypes. We denote such mapping as a transport plan $\Tmat$, obtained using Optimal Transport~\cite{ambrosio2008gradient,chen2020graph}.
Without loss of generality, we denote $\zv$ as the projected features for either image or text and optimize the following objective,

\begin{equation}
\label{eq:ot}
\small
\mathcal{L}_{\text{ot}} = \min_{\Tmat \in \Pi(\rvu,\rvv)}\sum^N_{i=1}\sum^K_{j=1}\Tmat_{ij} \cdot d(\zv^m_i,\cv_j) = \min_{\Tmat \in \Pi(\rvu,\rvv)} \,\, \langle \Tmat, \Dmat \rangle \,,
\end{equation}
where $\Pi(\rvu,\rvv) = \{ \Tmat \in \sR_+^{N\times K} | \Tmat\mathbf{1}_K=\frac{1}{N}\mathbf{1}_N, \Tmat^\top\mathbf{1}_N=\frac{1}{K}\mathbf{1}_K \} $, $\mathbf{1}_N$ denotes an $N$-dimensional all-one vector. $\Dmat$ is the cost matrix given by $\Dmat_{ij}=d(\zv^m_i,\cv_j)$ ($d(\cdot,\cdot) = 1 - \cos(\cdot,\cdot)$) and $\langle \Tmat, \Dmat \rangle=\text{Tr}(\Tmat^\top \Dmat)$ represents the Frobenius dot-product. We use Tg and Tf for the optimal transport plan for image and text in Algorithm~\ref{algo:frw-algo}, and $1-I2P$ corresponds to the cost matrix $\Dmat$ for image modality. It's similar for text. 

To solve for the optimal transport plan, we adopt an iterative algorithm shown in Algorithm \ref{alg:ipot}. It takes normalized feature matrix $\Zmat$, codebook $\Cmat$ as input and output an optimal tranpsort plan $\Tmat$. Internally, the algorithm tries to minimize the optimal transport (OT) distance, optimized to pick similar $\cv_j, j\in[1, \ldots,K]$ for each $\zv_i$ based on score $\Tmat[i,:]$ ($i^{th}$ row of $\Tmat$). In other words, $\Tmat$ can be viewed as a distance metric between prototypes and features.
When solved, OT yields a sparse solution $\Tmat^*$ containing at most $(2r-1)$ ($r=\max(N, K$) non-zero elements, leading to a robust and meaningful alignment~\cite{de2011optimal}.

In the codebook loss that we are going to formulate, $\Tmat$ will be used as ground-truth signals to guide the feature-to-prototype alignment. 
We use cross entropy loss and adopt a teacher-student distillation approach to construct the loss for optimizing the codebook as well as the feature encoders, 
\begin{align}
    \small
    \label{eq: codebook align loss}
    \Lcal_\text{t2p}(\Zmat_t, \Cmat, \Tmat_{i2p}) &= H(\Pmat_{t2p}, \Tmat_{i2p}), \nonumber\\ \small
    \Lcal_\text{i2p}(\Zmat_v, \Cmat, \Tmat_{t2p}) &= H(\Pmat_{i2p}, \Tmat_{t2p}),\\
    \Pmat_{t2p} = \textbf{SoftMax}(\Zmat_t \Cmat / \gamma)&, \Pmat_{i2p} = \textbf{SoftMax}(\Zmat_v \Cmat / \gamma) \nonumber
\end{align} 
where $\Pmat$ is the predicted metric calculated with the features from the student encoders while $\Tmat$ is calculated with features from the teacher encoders using Algorithm~\ref{alg:ipot}. The reason is that the teacher encoders are updated via exponential moving average, which helps avoid abrupt changes in codebook learning. 

\begin{figure}[t!]
    \centering
    \includegraphics[width=0.46\textwidth]{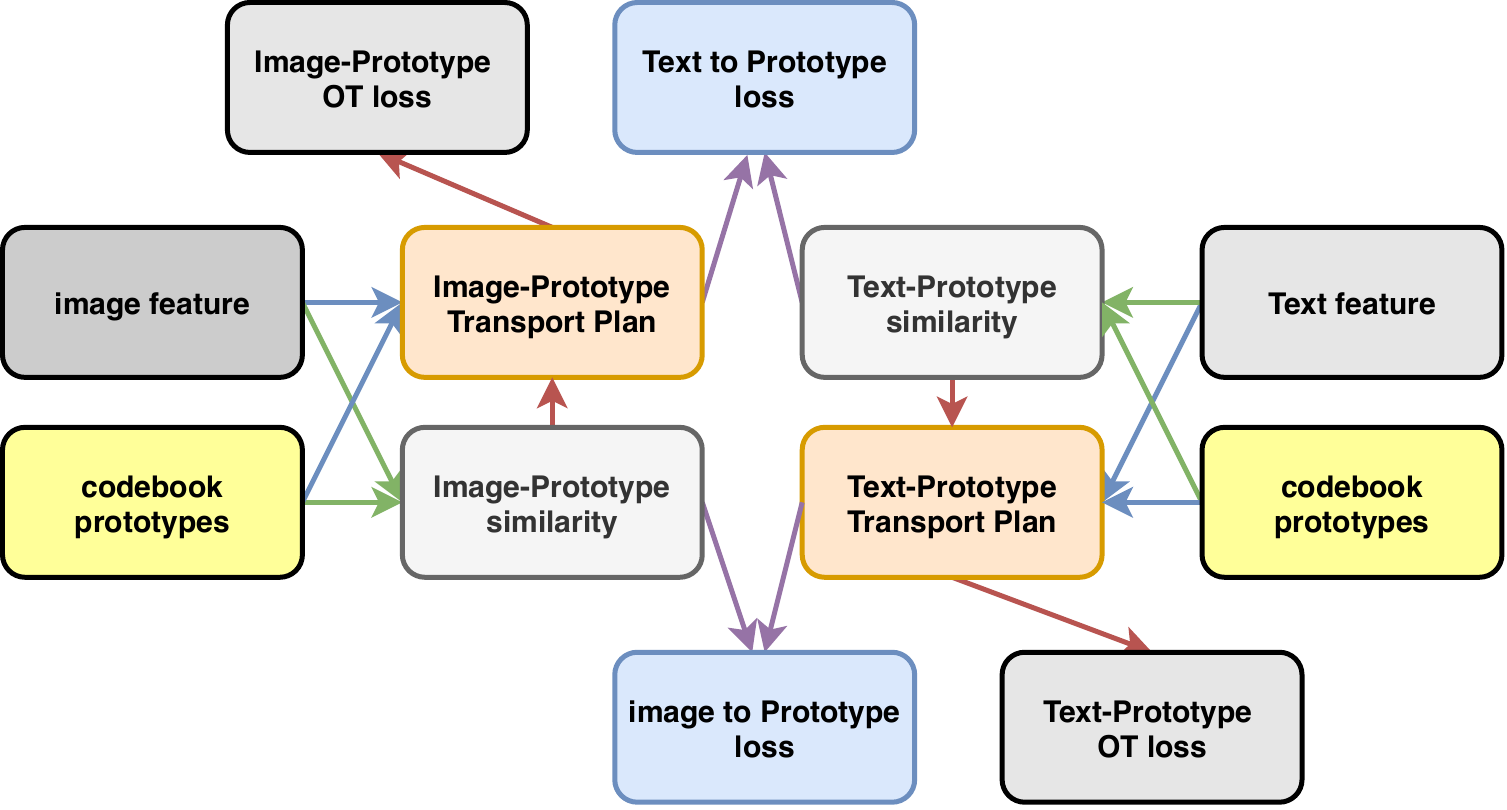}
    \caption{\small This is the diagram illustrating how to calculate four codebook losses. ``{\color{green} $\rightarrow$}": softmax operator. ``{\color{blue} $\rightarrow$}": IPOT algorithm. ``{\color{red} $\rightarrow$}": OT loss. ``{\color{purple} $\rightarrow$}": cross entropy.}
    \label{fig:codeloss}
    \vspace{-3mm}
\end{figure}

We additionally add a regularization term $ \Lcal_\text{ot}$. The overall loss for multimodal codebook learning is as follows,
\begin{equation} 
\begin{split}
    \Lcal_{\text{code}} &= \Lcal_{\text{ot}}(\Zmat^m_v, \Cmat) + \Lcal_{\text{ot}}(\Zmat^m_t, \Cmat) \\
    &+  \Lcal_\text{t2p}(\Zmat_t, \Cmat, \Tmat_{t2p}) + \Lcal_\text{i2p}(\Zmat_v, \Cmat, \Tmat_{i2p})
    \vspace{-5mm}
\end{split}
\end{equation}
As shown in Figure~\ref{fig:codeloss}, codebook acts as a bridge between the image and text modality, as both text to prototype loss ($\Lcal_\text{t2p}$) or image to prototype loss ($\Lcal_\text{i2p}$) chain features from both modalities. For example, Text to Prototype loss chains Image-Prototype Transport Plan and Text-Prototype Similarity and vice versa. More importantly, learning codebook allows contrasting features across modalities at the prototype level, i.e, feature distribution matching. When calculating the transport plan, we use the teacher features as they provide a more stable supervision signal to guide the learning of the student. The calculated losses will be backpropagated to update both the codebook and student encoders. 

\subsection{Teacher-student Distillation Learning} \label{sec: itc} 
This loss is designed to align the features from two uni-modal encoders, which is inspired by the recent success of SSL learning~\cite{he2020momentum,caron2021emerging}. Our motivation is that image and text can be treated as two ``views'' of the same entity, and we adopt a teacher-student distillation paradigm to align them. 
Since the raw feature directly from unimodal encoders are in different feature spaces, we learn a joint embedding space of dimension $256$, $\zv_v \in \Rcal^{256}, \zv_t  \in \Rcal^{256}$ for image and text student features. 
Following~\cite{he2020momentum,li2021align}, we store features from the teacher encoders $\zv^m_v \in \Rcal^{256}, \zv^m_t \in \Rcal^{256}$ in memory queues $\Qmat_v$, $\Qmat_t$ for image and text respectively.

For a pair of image and text, we can calculate the cross-modal similarity and intra-modal similarity as follows:
\begin{align}
\label{eq: ita_sim}\footnotesize
    \pv_{t2i}(T)=\exp{\frac{\zv_t \zv^{m\top}_v}{\gamma}} / \sum_{\zv^{m'}_v\in \Qmat_v} \exp{\frac{\zv_t \zv^{m'\top}_v}{\gamma}} \nonumber\\
    \pv_{i2t}(I)=\exp{\frac{\zv_v \zv^{m\top}_t}{\gamma}} / \sum_{\zv^{m'}_t\in \Qmat_t} \exp{\frac{\zv_v \zv^{m'\top}_t}{\gamma}} \\
    \pv_{i2i}(I)=\exp{\frac{\zv_v \zv^{m\top}_v}{\gamma}} / \sum_{\zv^{m'}_v\in \Qmat_v} \exp{\frac{\zv_v \zv^{m'\top}_v}{\gamma}} \nonumber\\
    \pv_{t2t}(T)=\exp{\frac{\zv_t \zv^{m\top}_t}{\gamma}} / \sum_{\zv^{m'}_t\in \Qmat_t} \exp{\frac{\zv_t \zv^{m'\top}_t}{\gamma}} \nonumber
\end{align}
where pseudo image negatives for estimating $\pv_{t2i}(T)$ is sampled from the image queue $\Qmat_v$ and similarly for $\pv_{i2t}(I)$. In addition to~\cite{li2021align}, we also considered unimodal (intra) alignment. Intuitively, enhancing unimodal feature representation lays a better foundation for cross-modal alignment. 

To further smooth out the learning process, we use the features from the momentum teacher to provide the soft distillation target, $\yv_{i2t}, \yv_{t2i}, \yv_{t2t}, \yv_{i2i}$ (refer to Algorithm~\ref{algo:frw-algo} for details).
The loss for intra/cross-modal alignment is defined as, 
\begin{align} \small
\label{eq: itc}
    \Lcal_{ica} = \Ep_{I,T\sim \pv_\text{data}} &\ [H(\pv_{t2t}, \yv_{t2t}) + H(\pv_{i2i}, \yv_{i2i}) \nonumber \\
    & + H(\pv_{t2i}, \yv_{t2i}) + H(\pv_{i2t}, \yv_{i2t})]
\end{align}
where $H$ is cross entropy. This objective can also be viewed as knowledge distillation, between teacher encoders and student encoders from the same modality (i.e., $H(\pv_{t2t}, \yv_{t2t})$ and $ H(\pv_{i2i}, \yv_{i2i}$), as well as between teacher encoders and student encoders from different modality (i.e., $H(\pv_{t2i}, \yv_{t2i})$ and $H(\pv_{i2t}, \yv_{i2t}$)). Parameters for the teacher encoder is an exponential moving average of the student, detached from gradient update. We adopt momentum update similar to~\cite{he2020momentum} to update the teacher encoders:
\begin{equation}
    f_t = \alpha f_t + (1-\alpha) f_s, g_t = \alpha g_t + (1-\alpha) g_s
\end{equation}
$\alpha$ is the momentum parameter. In practice, we set $\alpha=0.995$, in order to smoothly update teacher encoders.

\subsection{Self-supervised Pre-training}
\label{sec:pretrain}
In this section, we will first introduce two commonly used objectives for multimodal training frameworks: (i) masked language modeling loss (MLM) and (ii) image-text matching (ITM) on the multimodal encoder. Then we discuss how codebook and teacher-student distillation components are integrated. We denote the image and text features extracted by student network as $\{v_{cls}, v_1, ..., v_m\}$ and $\{t_{cls}, t_1, ..., t_n\}$, respectively.
Specifically, $v_{cls}$ is the image [CLS] token, $\{v_1, ..., v_m\}$ are image patch embeddings.
Similarly, $t_{cls}$ indicate the text [CLS] token, $\{t_1, ..., t_n\}$ are word embeddings.

\subsubsection{Image-Text Matching (ITM) Loss}
To fuse vision and language representations, we adopt ITM that is widely used in modern V\&L frameworks. Given an arbitrary pair of image and text, ITM predicts whether they are aligned (positive pairs) or not (negative pairs). This procedure can be formulated as a binary classification problem.

Specifically, [CLS] token from the fusion encoder is used as the joint representation of the image-text pair. ITM head is a fully connected layer to predict the matching probability $\pv_{\text{itm}}$. 
We assume that each image-text pair $(I_i, T_i)$ sampled from the pre-training datasets is a positive example and construct negative examples through the following strategy:
For each image $I_i$ within the batch, we sample one negative text $T_j$ from the same batch based on the contrastive similarity distribution. So that text that is more similar to this image will have a higher chance to get sampled. Similarly, one hard negative image will be sampled for each text $T_i$.
We denote $y_{\text{itm}}$ as the ground-truth labels indicating whether the image-text pair is positive or negative.
\begin{equation}
    \Lcal_{\text{itm}} = \Ep_{I,T\sim \pv_\text{data}} H(\pv_{\text{itm}}, \yv_{\text{itm}})
\end{equation}
where $H$ is the cross entropy operator.

\subsubsection{Masked Language Modeling (MLM) Loss}
We follow the design of MLM loss from BERT \cite{devlin2018bert}, which aims to predict the ground-truth labels of masked text tokens $y_{\text{mlm}}$.
Specifically, we randomly mask out 15\% of input text tokens, those masked tokens are replaced with special token [MASK].
Different from BERT, our MLM loss is conditioned on both surrounding text tokens and image representations.
Assume the predicted token probability is $\pv_{\text{mlm}}$, we construct the loss objective as follows, 
\begin{equation}
    \Lcal_{\text{mlm}} = \Ep_{I,\hat{T}\sim \pv_\text{data}} H(p_{\text{mlm}}, \yv_{\text{mlm}})
\end{equation}
where $\hat{T}$ is the text token sequence after masking.

\subsection{Summary}
We simultaneously optimize the codebook and the student encoders within the framework in an end-to-end manner, employing the losses discussed in previous sections as follows, 
\begin{equation}
    \Lcal_{\text{final}} = \Lcal_{\text{mlm}} + \Lcal_{\text{itm}} + \Lcal_{\text{ica}} + \Lcal_{\text{code}}
\end{equation}
among which MLM and ITM loss have been widely used in many V\&L methods particularly those ``early-fusion" frameworks. The ica loss is the main objective function for ``late-fusion" V\&L frameworks. CODIS combines the merits of both ``early-fusion'' and ``late-fusion'' approaches, by explicitly learning alignment along with fusion. 

Intra-cross alignment ($\Lcal_{\text{ica}}$) loss described in Section~\ref{sec: itc} can be viewed as an instance-to-instance alignment loss, similar to the one in~\cite{li2021align}. The difference is we consider both intra and cross modal alignment. We assume that a stronger unimodal representation can lay a solid foundation for cross-modal representation. Empirical evidence is provided in Section~\ref{sec:ablation}. The codebook loss ($\Lcal_{\text{code}}$) designed in Section~\ref{sec: code} measures the the distance between the transport plan and similarity matrix. It contrasts features at the prototype level and can be interpreted as distance metric matching~\cite{caron2018deep,chen2020graph}. Combining these two help avoid prototype collapsing problem, as online prototype clustering requires careful tuning~\cite{caron2020unsupervised}. Finally, The supervision signals for both intra-cross alignment loss and codebook loss require features from the momentum teacher and we adopt a teacher-student distillation approach. This can be seen as a generalization of unimodal SSL into the multimodal setting, under the V\&L framework.


\section{Experiments}\label{experiments}
To evaluate our approach, we conduct extensive studies on commonly used benchmarks and present experimental comparisons against state-of-the-art V\&L methods as shown in this section. We follow previous experimental protocols \cite{chen2020uniter, li2021align} for fair comparisons. We use Conceptual Captions (CC3M) \cite{sharma2018conceptual}, Visual Genome (VG) \cite{krishna2017visual}, SBU Captions \cite{ordonez2011im2text} and  COCO \cite{lin2014microsoft} as the pre-training dataset in our study, where a total of 4.0M unique images and 5.1M image-text pairs are covered.
\begin{table*}
	\footnotesize
	\setlength\tabcolsep{8pt}
	\caption{Performance comparison of zero-shot image-text retrieval on MSCOCO and Flickr30K datasets.\vspace{-3mm}}
	\begin{center}
		\begin{tabular}{ccccccc|cccccc}
            \toprule 
                \multirow{3}{*}{Method} & \multicolumn{6}{c}{MSCOCO (5K)} & \multicolumn{6}{c}{Flickr30K (1K)} \\
                
                & \multicolumn{3}{c}{Text Retrieval} & \multicolumn{3}{c}{Image Retrieval} &
                \multicolumn{3}{c}{Text Retrieval} & \multicolumn{3}{c}{Image Retrieval} \\
            
                & R@1 & R@5 & R@10 & R@1 & R@5 & R@10
                & R@1 & R@5 & R@10 & R@1 & R@5 & R@10 \\
                \midrule
                ImageBERT \cite{qi2020imagebert} & 44.0 & 71.2 & 80.4 & 32.3 & 59.0 & 70.2 & 70.7 & 90.2 & 94.0 & 54.3 & 79.6 & 87.5  \\
                Unicoder-VL\cite{li2020unicoder}  &  - & - & - & - & - & - & 64.3 & 85.8 & 92.3 & 48.4 & 76.0 & 85.2  \\
                UNITER \cite{chen2020uniter} &  - & - & - & - & - & - &  80.7 & 95.7 & 98.0 & 66.2 & 88.4 & 92.9  \\
                ViLT \cite{kim2021vilt} & 56.5 & 82.6 & 89.6 &  40.4 & 70.0 & 81.1& 73.2 & 93.6 & 96.5 &  55.0 & 82.5 & 89.8  \\
                CLIP \cite{radford2021learning} & 58.4 & 81.5 & 88.1 & 37.8 & 62.4 & 72.2&  88.0 & 98.7 & 99.4 & 68.7 & 90.6 & 95.2  \\
                
                ALIGN \cite{jia2021scaling} &  58.6 & 83.0 & 89.7 & 45.6 & 69.8 & 78.6& 88.6 & 98.7 &  \textbf{99.7} & 75.7 & 93.8 &  \textbf{96.8}  \\
                
                ALBEF 4M \cite{li2021align} & 68.6 & 89.5 & 94.7 & 50.1 & 76.4 & 84.5& 90.5 & 98.8 &  \textbf{99.7} & 76.8 & 93.7 & 96.7  \\
                \rowcolor{Light}        
                \bf Ours  &  \textbf{71.5} &  \textbf{91.1} &  \textbf{95.5} &  \textbf{53.9} &  \textbf{79.5} &  \textbf{87.1} & \textbf{91.7} &  \textbf{99.3} & 99.8 &  \textbf{79.7}&  \textbf{94.8} & 97.3 \\
                
                \bottomrule
            \end{tabular}
	\end{center}
	\label{table:zero_shot}
\end{table*}
\newcommand{\xmark}{\ding{55}}
\begin{table*}
	\footnotesize
	\setlength\tabcolsep{8pt}
	\caption{Performance comparison of fine-tuned image-text retrieval on MSCOCO and Flickr30K datasets.\vspace{-3mm}}
	\begin{center}
		\begin{tabular}{ccccccc|cccccc}
            \toprule 
                \multirow{3}{*}{Method} & \multicolumn{6}{c}{MSCOCO (5K)} & \multicolumn{6}{c}{Flickr30K (1K)} \\
                
                & \multicolumn{3}{c}{Text Retrieval} & \multicolumn{3}{c}{Image Retrieval} &
                \multicolumn{3}{c}{Text Retrieval} & \multicolumn{3}{c}{Image Retrieval} \\
            
                & R@1 & R@5 & R@10 & R@1 & R@5 & R@10
                & R@1 & R@5 & R@10 & R@1 & R@5 & R@10 \\
                \midrule
                ImageBERT \cite{qi2020imagebert} & 66.4 & 89.8 & 94.4 & 50.5 & 78.7 & 87.1&  87.0 & 97.6 & 99.2 & 73.1 & 92.6 & 96.0  \\    
            
                UNITER \cite{chen2020uniter} & 65.7 & 88.6 & 93.8 & 52.9 & 79.9 & 88.0  &  87.3 & 98.0 & 99.2 & 75.6 & 94.1 & 96.8 \\
                
                VILLA \cite{gan2020large} & - & - & - & - & - & - & 87.9 & 97.5 & 98.8 & 76.3 & 94.2 & 96.8  \\
                
                OSCAR \cite{li2020oscar} & 70.0 & 91.1 & 95.5 &  54.0 & 80.8 & 88.5 & - & - & - & - & - & -  \\
                
                ViLT \cite{kim2021vilt} & 61.5 & 86.3 & 92.7  & 42.7 & 72.9 & 83.1 & 83.5 & 96.7 & 98.6 &  64.4 & 88.7 & 93.8  \\
                
                UNIMO \cite{li2020unimo}  & - & - & - & - & - & - & 89.7 & 98.4 & 99.1 & 74.6 & 93.4 & 96.0 \\
                
                
                
                SOHO \cite{huang2021seeing} & 66.4 & 88.2 & 93.8 & 50.6 & 78.0 & 86.7 & 86.5 & 98.1 & 99.3 &  72.5 & 92.7 & 96.1  \\
                
                ALBEF 4M \cite{li2021align} &  73.1 & 91.4 & 96.0 & 56.8 & 81.5 & 89.2 & 94.3 & \textbf{99.4} & 99.8 & 82.8 & 96.7 & \textbf{98.4}  \\
                
                \rowcolor{Light}
                \bf Ours  & \textbf{75.3} & \textbf{92.6} & \textbf{96.6} & \textbf{58.7} & \textbf{82.8} & \textbf{89.7}&
                \textbf{95.1} & \textbf{99.4} & \textbf{99.9} & \textbf{83.3} & 96.1 & 97.8  \\
            
                \bottomrule
            \end{tabular}
	\end{center}
	\label{table:fine_tune}
\end{table*}

\subsection{Downstream Tasks}

\noindent\textbf{Image-Text Retrieval} consists of two tasks: (1) image as query and retrieve texts (TR); (2) text as query and retrieve images (IR). 
The pre-trained model is evaluated on MSCOCO \cite{lin2014microsoft} and Flickr30K \cite{plummer2015flickr30k}. For the zero-shot setting, the pre-trained model is directly evaluated on the test data without any further training. In particular, for zero-shot retrieval on Flickr30K, we follow the procedure proposed in \cite{li2021align} (zero-shot evaluating on Flickr with the model fine-tuned using MSCOCO).
For the fine-tuning setting, the pre-trained model is fine-tuned on the training data and evaluated on the validation/test data.

\noindent\textbf{Visual Question Answering (VQA) \cite{goyal2017making}} 
predicts the answer given an image and a question, which requires an understanding of vision, language and context.
We consider this task as a generation problem by finetuning an answer decoder to generate the answer from candidates as in \cite{li2021align}.

\noindent\textbf{Visual Reasoning (NLVR$^2$)}
The dataset~\cite{suhr2018corpus} contains 107,292 examples of human-written English sentences paired with web photographs. The task is to determine whether a natural language caption is true about a pair of photographs. We extend our model as \cite{li2021align} to take a text and two images as input. 

\noindent\textbf{Visual Entailment (SNLI-VE) \cite{xie2019visual}} 
predicts whether a given image entails a given text, which is formulated as a three-way classification problem (entailment, neutral, or contradictory) in our framework.


\subsubsection{Implementation Details}
We adopt ViT-B/16~\cite{dosovitskiy2020image} as our vision encoder. The text encoder uses BERT\textsubscript{base} with 12 layers.
We set queue size to be $65,536$, codebook size as $4000$ and moving average $\alpha=0.995$.
For the pre-training stage, the model is trained for 30 epochs with a batch size of 512.
We use mini-batch AdamW optimizer \cite{loshchilov2017decoupled} with a weight decay of 0.02.
The learning rate is initialized as $1e-5$ and warmed-up to $1e-4$ after 1,000 iterations. 
Then it's decreased with a cosine decay strategy to $1e-5$. All of our experiments were performed on $8$ NVIDIA A100 GPUs.
The image input is randomly cropped and resized to 256$\times$256, before RandAugment \cite{cubuk2020randaugment} is applied. 
During fine-tuning, the image resolution is increased to 384$\times$384 for fair comparison with existing approaches~\cite{li2021align}.

\subsection{Evaluation on Image-Text Retrieval}
\label{sec:zero-shot1}
For the image-text retrieval tasks, we conduct two different scenarios for evaluation: ``zero-shot" retrieval task and ``after-finetuning" retrieval task, following the setting in~\cite{li2021align,chen2020uniter,li2020oscar}. We compare with both early-fusion methods such as~\cite{chen2020uniter,li2020oscar,kim2021vilt} and late-fusion methods such as~\cite{radford2018improving,jia2021scaling}. ALBEF~\cite{li2021align} is an hybrid approach that also performs feature alignment along with fusion. Results in Table \ref{table:zero_shot} and \ref{table:fine_tune} show consistent improvements of our approach against prior state-of-the-arts.

\textbf{``Zero-shot"}: As shown from Table \ref{table:zero_shot}, CODIS outperforms existing baselines with a clear margin across the two datasets, for both image and text retrieval tasks, especially at R@1. Compared to the best-performing early-fusion approach~\cite{chen2020uniter}, we obtain a margin of $11.0\%$/$13.5\%$ TR/IR in terms of R@1 on Flickr30K. When compared to highest late-fusion approach~\cite{jia2021scaling}, there's an increase of $12.9\%$/$8.3\%$ TR/IR in R@1 on MSCOCO and a boost of $3.1\%$/$4.0\%$ TR/IR in R@1 on Flickr30K, despite the fact that ALIGN~\cite{jia2021scaling} uses 1.8B data in training (approx. 360$\times$ more image-text pairs than our model). Our approach also outperforms ALBEF 4M~\cite{li2021align} with a clear margin of 2.9\%/3.8\% R@1 for TR/IR on MSCOCO and 1.2\%/2.9\% in terms of R@1 for TR/IR on Flickr30K, revealing that our model can further benefit from codebook representation learning. 

\textbf{``After-finetuning"}: This task showcases the ability of V\&L pretraining via transfer learning. For small datasets such as Flickr30K, performance gap tends to reduce as the model converges. However, our approach still achieves the best result in most of the metrics and the largest margins occur for R@1, especially on MSCOCO. Compared against the closest performing method ALBEF~\cite{li2021align}, CODIS obtains an improvement of $2.2\%/1.9\%$ TR/IR in R@1 on MSCOCO, providing evidence to the effectiveness of CODIS for transfer learning. 

\begin{table}[!t]
    \aboverulesep = 0.48mm
    \belowrulesep = 0.48mm
    \small
    \footnotesize
	\centering	
    \caption
	{
		Comparison with variety of state-of-the-art methods on downstream vision-language tasks: VQA, NVLR$^2$, SNLI-VE.
	}
	\resizebox{0.5\textwidth}{!}{%
	\begin{tabular}	{l   |  c  c  c  c  c  c  }
		\toprule	 	
	 \multirow{2}{*}{Method} & \multicolumn{2}{c}{VQA} & \multicolumn{2}{c}{NLVR$^2$} & \multicolumn{2}{c}{SNLI-VE} \\
	  & test-dev & test-std & dev & test-P & val & test\\
	  \midrule
	  VisualBERT~\cite{li2019visualbert} & 70.80 & 71.00 & 67.40 & 67.00 & - & - \\
	  LXMERT~\cite{tan2019lxmert}  & 72.42 & 72.54 & 74.90 & 74.50 & - & - \\
	  12-in-1~\cite{lu202012} & 73.15 & - & - & 78.87 & - & 76.95 \\
	  UNITER~\cite{chen2020uniter} & 72.70 & 72.91 & 77.18 & 77.85 & 78.59 & 78.28 \\
	   ViLT~\cite{kim2021vilt}  & 70.94 & - & 75.24 & 76.21& - & - \\
	   OSCAR~\cite{li2020oscar} &  73.16 & 73.44 & 78.07 & 78.36 & - & - \\
	   VILLA~\cite{gan2020large} & 73.59 & 73.67 & 78.39 & 79.30 & 79.47 & 79.03 \\
	  ALBEF 4M\cite{li2021align}  & 74.54 & 74.70 & 80.24 & 80.50 & 80.14 & 80.30 \\
	  \midrule
	  \rowcolor{Light}
	   \bf Ours & \textbf{74.86} & \textbf{74.97} & \textbf{80.50} &  \textbf{80.84}& \textbf{80.47} &  \textbf{80.40} \\
		\bottomrule
	\end{tabular}
 	}
	\label{tbl:vqa_nlvr_ve}
	\vspace{-3.5ex}

\end{table}		
\subsection{Evaluation on VQA, NLVR and VE}
Following previous approaches~\cite{chen2020uniter,li2021align}, we further report performances of CODIS on various other vision-language tasks such as VQA, NLVR and VE. It's worth noting that some results are not directly comparable as~\cite{chen2020uniter} additionally uses out-of-domain data, ~\cite{li2020oscar} leverages additional object tags and~\cite{gan2020large} with adversarial data augmentation. Nevertheless, we observe consistent improvement of our method on all tasks across different datasets in Table \ref{tbl:vqa_nlvr_ve}.

\subsection{Ablation Study} \label{sec:ablation}
In this section, we do ablation studies on the performance of our approach with different variants of CODIS. To get a clear understanding about the effects of each component, we perform comparisons under the zero-shot setting without any finetuning. Note that the setting here for Flickr30K is different than the one in Section~\ref{sec:zero-shot1}, as the latter reports numbers based on the finetuned model on MSCOCO (5K). Refer to ~\cite{chen2020uniter} for more details. 

\begin{table}[h]
	\scriptsize
	\vspace{-0.1in}
	\setlength\tabcolsep{5pt}
	\caption{Efficiency of our approach under limited pretraining regime using only MSCOCO.}
	\vspace{-3mm}
	\begin{center}
		\begin{tabular}{l|ccc|ccc}
                & TR@1 & TR@5 & TR@10 & IR@1 & IR@5 & IR@10 \\
                \midrule
                ALBEF  & 55.70 & 81.92 & 88.78 & 41.08 & 69.01 & 78.86 \\
                \midrule 
                0.5x codebook & 58.66 & 83.9 & 90.64 & 43.74 &72.10 & 81.58  \\
                2.0x codebook & 59.02 & 84.46 & 91.06 & 43.62 & 71.69 & 81.12  \\
                \midrule
                3K codewords & 58.96 & 84.28 & 90.98 & 44.66 & 72.31 & 81.68 \\
                500 codewords & 55.52 & 81.68 & 89.28 & 41.53 & 68.75 & 78.43 \\
                \midrule
                \textbf{Ours} & 59.38 & 84.04 & 91.20 & 44.71 & 72.63 & 81.69 \\
            \end{tabular}
	\end{center}
	\label{table:small}
	\vspace{-0.3in}
\end{table}

Results are summarized in Table \ref{table:ablation}. By removing the effect of codebook, we provide two baselines that perform alignment at the instance level, namely (a) cross-modal alignment only and (b) intra + cross alignment. The former is an equivalent of ALBEF~\cite{li2021align}, as both consider only alignment across modalities. The performances consistently increase for all R@1 TR/IR metrics (+1.26\%/+0.42\% on in R@1 for TR/IR on MSCOCO and +0.9\%/+1.52\% in R@1 for TR/IR on Flickr) by involving intra-modal alignment, i.e., enhancing unimodal representations.

\begin{table*}
	\footnotesize
	\setlength\tabcolsep{8pt}
	\caption{Performance comparison of zero-shot image-text retrieval on Flickr30K and COCO datasets for ablation study.	\vspace{-3mm}\label{table:ablation}}
	\begin{center}
	\resizebox{\linewidth}{!}{
		\begin{tabular}{lgccgcc|gccgcc}
            \toprule 
                \multirow{3}{*}{Objective functions} &
                \multicolumn{6}{c}{MSCOCO (5K)} & \multicolumn{6}{c}{Flickr30K (1K)} \\
                
                & \multicolumn{3}{c}{Text Retrieval} & \multicolumn{3}{c}{Image Retrieval} &
                \multicolumn{3}{c}{Text Retrieval} & \multicolumn{3}{c}{Text Retrieval} \\
            
                & R@1 & R@5 & R@10 & R@1 & R@5 & R@10
                & R@1 & R@5 & R@10 & R@1 & R@5 & R@10 \\
                \midrule
                a: MLM+ITM+ITC (cross align)  & 68.60 & 89.50 & 94.70 & 50.10 & 76.40 & 84.50 & 84.90 & 97.20 & 99.00 & 68.18 & 88.58 & 93.02  \\
                b: MLM+ITM+ITC (intra + cross) &  69.86 & 89.48 & 94.42 & 50.52 & 77.02 & 85.17 & 85.80 & 96.80 & 98.10 & 69.70 & 89.60 & 93.48  \\
                \midrule
                a + codebook (teacher feature)  &  70.74 & 89.54 & 94.88 &  51.39 & 77.86 & 85.60  &  86.00 & 97.00 & 98.20 &  70.18 & 90.66 & 94.44 \\
                b + codebook (student feature)   &  71.12 & 89.62 & 94.78 &  51.40 & 77.42 & 85.53  &  86.30 & 96.90 & 98.30 &  70.34 & 90.00 & 93.84\\
                b + codebook (teacher feature) &  \textbf{71.10} &  \textbf{90.60} &  \textbf{95.10} &  \textbf{52.10} &  \textbf{78.00} &  \textbf{85.90}  &  \bf 86.70 & \bf 97.30 & 98.70 &  \bf 71.40 & \bf 90.82 & \bf 94.62  \\
                \bottomrule
            \end{tabular}
            }
	\end{center}

\end{table*}
\begin{figure*}[!h]
  \centering
   \includegraphics[width=0.80\linewidth]{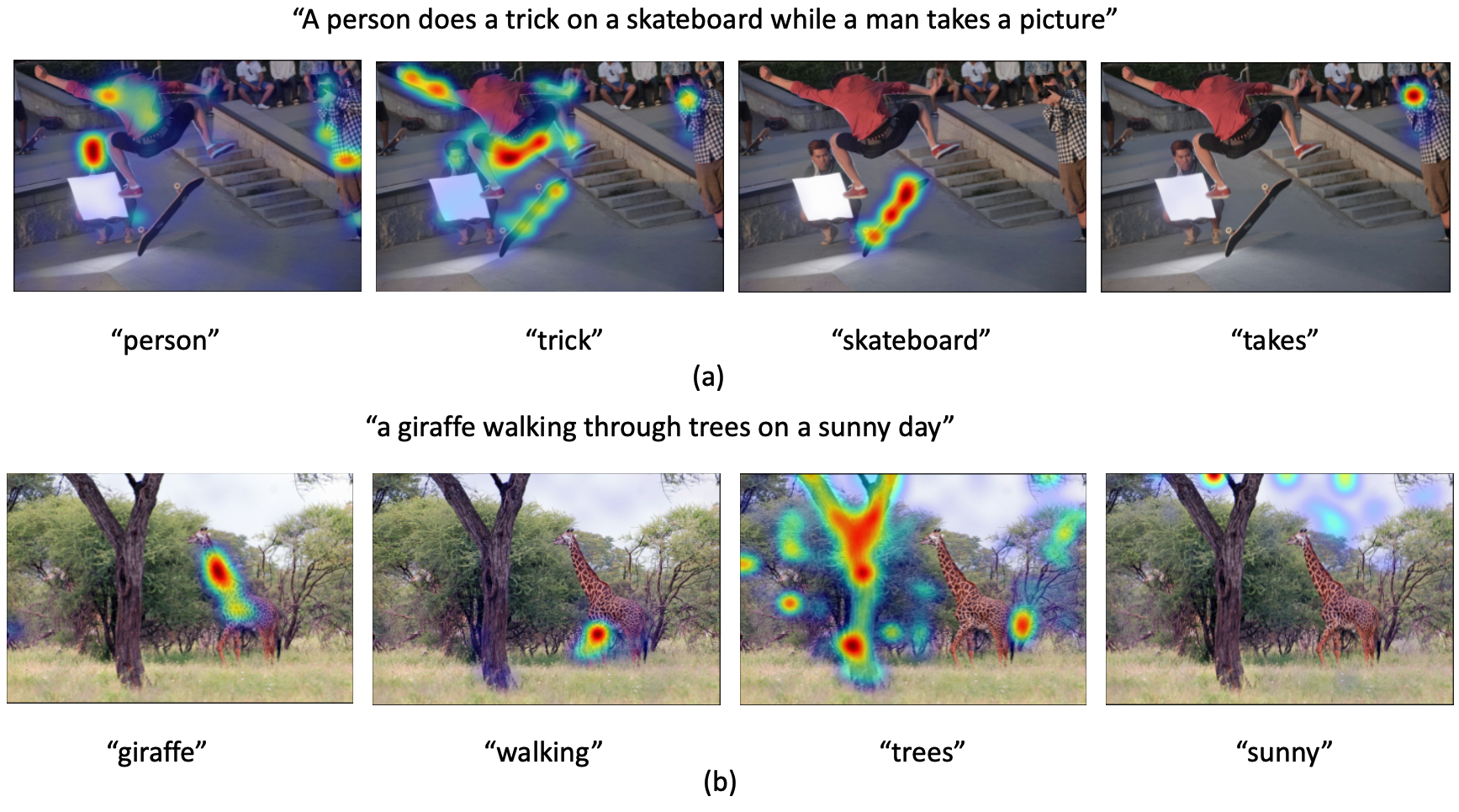}
   \vspace{-2mm}\caption{Grad-CAM visualization on the cross-attention maps corresponding to individual words   }
   \label{fig:visual}
\end{figure*}

We observe a consistent improvement over the two baselines when codebook is considered. In this genre, we provide three variants of CODIS designs. The 1st and 3rd row compare the effects of intra-modal alignment whereas the 2nd and 3rd row studies the effects of using student and teacher features for computing the codebook loss. This experiment also serves to support the validity by combining teacher-student distillation with codebook representation learning. Combining the two contributions, CODIS improves the first baseline by a clear margin of $2.5\%/2.0\%$ in R@1 for TR/IR on MSCOCO and $1.8\%/3.22\%$ absolute R@1 for TR/IR on Flickr. 

To further investigate the efficiency of our approach, we provide ablations on different codebook loss weights and codebook sizes on MSCOCO test when pretrained with MSCOCO train split under the zero-shot setting in Table~\ref{table:small}. 

\subsection{Cross-attention visualization}
We visualize the cross-attention maps using Grad-CAM\cite{selvaraju2017grad} to provide qualitative assessment of CODIS. Figure~\ref{fig:visual} shows that CODIS is able to associate language with ``regions of interest'' by attending to meaningful objects and locations, visually reflecting the quality of our model in multimodal alignment. For example, in the first row, the model attends to all men when word ``person'' is given, while for words such as ``tricks'' and ``takes'', the model performs surprisingly well, by ``focusing'' exclusively on the related persons. In the second example, we choose a scene where multiple correspondences exist (e.g., trees and sunny day). The model seems to allocate more attention to trees closest to the camera and can differentiate trees from grass. It's interesting to observe that the model switches its ``attention'' from the upper-body of the giraffe to its feet when the word changes from ``giraffe'' to ``walking'', demonstrating the model's capability in understanding the semantic relations between image and text.

\section{Conclusion and Future Work}\label{sec:conclusion}
Vision and language pretraining is attracting growing attention of the computer vision community and has exhibited great potential across a diversity of vision-language downstream tasks. One of the keys to the success of V\&L is to improve multimodal alignment. In this paper, we propose multimodal alignment using representation codebook, which acts as a medium between the modalities. We also make a connection between self-supervised learning and V\&L pretraining, by generalizing teacher-student distillation learning to the multimodal setting under the V\&L framework. Our work is a step toward more principled multimodal alignment. We hope to inspire more works in this direction. 

\paragraph{Acknowledgement}
The authors would like to thank Chenyang Tao for helpful comments on CODIS experiments. 

{\small
\bibliographystyle{ieee_fullname}
\bibliography{egbib}
}

\end{document}